\documentclass[10pt,twoside]{article}

\usepackage{times}
\usepackage[utf8]{inputenc}
\usepackage[T1]{fontenc}
\usepackage{graphicx}
\usepackage{comment}
\usepackage{todonotes}
\usepackage{url}
\usepackage{multirow}

\usepackage{taln2019}
\usepackage[frenchb]{babel}

\title{Exploring sentence informativeness}

\author{Syrielle Montariol\up{1, 2}  \quad Aina Gar\'i Soler\up{1}  \quad Alexandre Allauzen\up{1}\\
  {\small
    (1) LIMSI, CNRS, Univ. Paris-Sud, Univ. Paris-Saclay, F-91405 Orsay, France \\ 
    (2) Société Générale, 17 Cours Valmy 92043 Puteaux, France \\ 
    \texttt{
       syrielle.montariol@limsi.fr, aina.gari@limsi.fr, alexandre.allauzen@limsi.fr \\
}}}

\begin{document}
\maketitle

\resume{Explorer l'informativité d'une phrase}{
  Nous présentons ici une exploration préliminaire du concept d'\textit{informativité} --la quantité d'information qu'une phrase fournit sur l'un des mots qui le compose-- et ses usages potentiels pour l'apprentissage de plongements de mots robustes à partir de données en faible quantité. Une mesure d'informativité est prédite à partir d'algorithmes de classification de phrases, que nous comparons à une série de phrases annotées manuellement. Nous concluons que ces deux mesures correspondent à des notions différentes d'informativité. Néanmoins, nos expériences montrent que la prédiction extraite de la classification a un impact sur la qualité des plongements de mots lors de l'apprentissage.
}

\abstract{}{
This study is a preliminary exploration of the concept of \textit{informativeness} --how much information a sentence gives about a word it contains-- and its potential benefits to building quality word representations from scarce data. We propose several sentence-level classifiers to predict informativeness, and we perform a manual annotation on a set of sentences. We conclude that these two measures correspond to different notions of informativeness. However, our experiments show that using the classifiers' predictions to train word embeddings has an impact on embedding quality.
}

\motsClefs
  {Informativité, Plongements de mots, Classification de phrases, Labellisation}
  {Informativeness, Word embeddings, Sentence classification, Data annotation}

\section{Introduction}

Building robust and high-quality word representations is a key step for most NLP tasks. The quality mostly relies on the training corpus (its size and relevance), along with the training criterion and strategy.
Some noisy sentences might for instance damage the quality of embeddings, 
while many sentences do not contribute significantly in improving the representation of a word's meaning (see Table \ref{sentEx} for an example). Therefore, it is advised to estimate embeddings on corpora as large as possible, but this induces many drawbacks: such corpora are difficult to collect for many domains or languages, and the training time increases linearly with the corpus size. Many pre-trained embeddings do exist, built on huge corpora, but they cannot be used in domain-specific tasks.

Consequently, a criterion for sentence selection could be of great help in the context of low-quality or low-volume corpora.
The concept at stake is the \textit{informativeness} of a sentence towards a word. This criterion can be defined as follows: 
\emph{a sentence is informative with respect to one of its words if a person ignoring this word can correctly infer its meaning from this sentence to some extent.}
The sentences from the training corpora can be selected depending on their informativeness during embeddings' training. To this end, the informativeness of a sentence needs to be defined and modelled. 

The contribution of this paper is threefold. First, we propose a method to automatically get an \textit{artificial} informativeness measure of any sentence towards a target word, using sentence classification. Then, we design a labelling process in terms of informativeness scores to evaluate our models' results. Finally, we give preliminary results on the use of the artificial informativeness measure for the training of word embeddings.

\begin{table}[!ht]
\centering
\begin{tabular}{|l|l|}
\hline
 A: \textit{I went to the ***  } & C: \textit{It was good *** in those terms.}\\                
 B: \textit{I went to the *** to withdraw money} & D: \textit{The aquarium did a blood ***} \\
 & \textit{to determine their gender.} \\
 \hline
\end{tabular}
\caption{Examples of informativeness disparity between two sentences with respect to a common word. A and B are made-up examples; C and D are Gigaword \citep{napoles2012annotated} sentences with the word \textit{test}.}
\label{sentEx}
\end{table}

\section{Related work}
\label{sec:related}


While dealing with low-resource languages and specific domains has received much attention in the NLP community, to the best of our knowledge, this is the first attempt to investigate the potential impact of the notion of informativeness on building meaning representations from limited data.
Several attempts at improving or adapting word embeddings to restricted tasks and languages involve making use of morphological information \citep{luong2013better}; fine-tuning pre-trained, global-purpose embeddings on a restricted domain \citep{komiya2018investigating, newman2018embeddingtransfer} 
; refining them with the help of semantic resources \citep{faruqui2015retrofitting}, known as retrofitting; or using attention mechanisms on contexts to better represent rare words \citep{schick2019attentive}.

\citet{lai2016generate} show that, when training word embeddings, using an in-domain corpus specialized in a task is better than having a large, mixed-domain corpus, which can lead to a decrease in performance.

The closest study to the present one is that of \citet{Baroni-highrisk}. The authors argue that just as humans do not need many examples to learn the meaning of a word, the word2vec architecture can be adapted to learn a competitive representation of a word from 2-6 occurrences only. This can be critical to build an embedding for rare words or in situations of scarce data. Their experiments involve both Wikipedia definitions --in principle, maximally informative from a human perspective-- and naturally-occurring sentences. Given the better performance of the former, they observe that accounting for the informativeness of a sentence can be useful to learn good representations.

A line of work related to informativeness concerns the automatic extraction of sentence examples for lexicographic resources or knowledge bases. \citet{kilgarriff2008gdex} restrict their search with different linguistic criteria designed to help a reader grasp the meaning of a word more easily, such as sentence length and the presence or absence of rare words, pronouns, or typical collocations. Another approach for finding \textit{knowledge-rich contexts}, as coined by \citet{meyer2001extracting}, consists in finding sentences with knowledge patterns, that is, linguistic expressions that describe the semantic relations of a word, such as "is a kind of" to denote hypernymy \citep{barriere2004knowledge}.

\section{Sentence classification for informativeness prediction}
\label{sec:classif}
According to our definition, if a sentence is very informative with respect to one of its words, a reader who does not know that word can infer its meaning from the sentential context. Intuitively, in such a sentence, a person who knows the word should be able to predict it with high accuracy.
We create a task based on this intuition: for a given target word, a model gets as input sentences with a blank at the target position. The model must learn to distinguish sentences that originally contained the target word from those which did not. 
For instance in Table \ref{sentEx}, a model that distinguishes \textit{bank} sentences from \textit{non-bank} sentences should have a higher confidence that the more informative sentence B is a \textit{bank}-sentence. 
For each sentence, the classifier outputs a probability of belonging to the \textit{bank}-sentences class.
Our hypothesis is that this probability is an indicator of the informativeness of the sentence with respect to the word \textit{bank}.

\subsection{Selection of distractors}

In order to make the classification task challenging enough for the models to be effective, we introduce \textit{distractors}: words that share contexts with the target word and could therefore deceive the model.
In the first example of Table \ref{sentEx}, a good distractor would be a word that, like \textit{bank}, indicates a place (\textit{supermarket}, for example).

For each target word, 10 distractors are selected following 
\citet{hill2016automatic}'s work. 
First, we collect 3, 4 and 5-grams which include the target word\footnote{With a minimum frequency of 40, 20 and 5 for 3, 4 and 5-grams, respectively.}
from the 1 million most frequent n-grams in the Corpus of Contemporary American English (COCA).\footnote{Available at \url{https://www.ngrams.info/}} We then search, in the COCA corpus, for words that appear in the extracted n-grams with the same part of speech and position. From this list of potential distractors, words that are synonyms, hyponyms or hypernyms of the target word according to WordNet \citep{Fellbaum1998} are removed. Finally, only candidates that have the same or higher frequency than the target word are kept, as calculated from Google unigrams \citep{brants2006web}. Ten of the remaining words are randomly chosen to be the target word's distractors.

\subsection{Classification algorithms} \label{ssec:classif}

The first classifier relies on a context2vec (c2v) model \citep{melamud2016c2v} pre-trained on the ukWaC corpus \citep{baroni2009wacky}. c2v embeddings are trained with a slot-filling objective and can compute comparable embedding representations of sentential contexts with a blank slot as well as of individual words, where context vectors have a high similarity with those of appropriate fillers. We experiment with logistic regression and a feed-forward neural network with two hidden layers using as input c2v context vectors with a slot at the target word's position. 

We compare this to  a logistic regression classifier that relies on 3 linguistic-based features to discriminate sentences. Concretely, we use a 3-gram language model\footnote{Available at  \url{http://www.keithv.com/software/giga/} (NVP, 64K words)} and the aforementioned c2v model. For the language model feature, the blank slot is successively replaced with each distractor of a target word. Then, the probability of every resulting sentence is computed. The feature used is the proportion of distractors that, according to the language model, have a higher probability than the target word of filling the slot. The c2v features are the cosine similarity between the target word and the context representation, and the average cosine similarity of every distractor with the context representation.

Several other classification algorithms from the literature were also tested. One with a high accuracy and very low computation time is the FastText classifier \cite{fastextclassif}. It relies on logistic regression; the sentences are represented with averaging the bag-of-ngrams representation of their words. We put a \textit{"TARGET"} token in place of the target word, a \textit{"NUMBER"} token in place of a number, and --as in the previous models-- keep stop words.




\section{Labelling data}
\label{sec:labelling}
Given a sentence, the classification algorithm outputs a probability of belonging to the class of the target word. According to our hypothesis, this method gives us a measure of informativeness for each pair [\textit{sentence, target word}]. In order to evaluate this measure, two annotators label a set of sentences. Then, we perform an inter-annotator agreement study to select the best methodology and scoring scale. 

For a first agreement, the two annotators both label 150 sentences randomly extracted from a portion of the Annotated English Gigaword corpus \footnote{Released by the Linguistic Data Consortium, see \url{https://catalog.ldc.upenn.edu/LDC2012T21}} \citep{napoles2012annotated}. 
In each sentence, a target word is randomly masked. It has to be a noun, verb, adjective or adverb, excluding proper nouns and auxiliary verbs. The annotators give two scores to each sentence: $info1$ before seeing the masked word (from 1 to 10, indicating how much they can guess about the target word given the sentence) and $info2$ after seeing this word (from 1 to 10, expressing to what extent they expected seeing the true word).\\
The Spearman correlations between both annotators' scores are relatively low (mean correlation = 0.29 for $info1$ and 0.37 for $info2$), especially for adverbs (correlation = 0.12 for $info2$). Overall, the annotators' remarks show that these measures are very subjective.

Consequently a second labelling agreement is designed, relying on a more explicit measure : $info3$. The range of scores is reduced to be from 1 to 5, and precise scoring guidelines are designed to ensure a common interpretation of scores across annotators. The $info3$ measure answers the question: \textit{How much information does this sentence give about the meaning of the target word?}

\begin{enumerate}
    \item The sentence gives no clue about the target word (e.g \textit{I have a ...})
    \item The sentence has at least one element (e.g. \textit{I went to the ....})
    \item The sentence gives some clues about the concept but not very specific (\textit{I went to the ... to speak with the manager.})
    \item The sentence gives a lot of information about the word, but not enough to define it. (\textit{I went to the ... to open a savings account.})
    \item From the sentence, I would be able to write a definition of the word, or this is the only word that could fit here. (\textit{I went to the ... to withdraw money, exchange dollars and ask for a loan.})
\end{enumerate}


Adverbs are excluded due to their low inter-annotator agreement. Two sentences for each of 50 selected target words are extracted from the same corpus, and manually annotated the same way as for $info2$: the annotators know which word is the target during labeling. \\
The global Spearman correlation is 0.331. The annotators rarely disagree on extreme scores, rather on medium scores (between 2 and 4). However, the correlation is very low for adjectives (0.080). \\
To sum up, the second labelling agreement gives a more objective scale, making the annotators usually have close scores, and agreeing on extreme scores. Thus, we keep this process for the rest of the labeling. 20 words are selected out of the 50, excluding adjectives: \textit{call, go, range, carry, charge, coach, hold, return, check, investigator, shot, education, paper, side, figure, post, tell, fire, put, test}. The annotators label together five more sentences for each of the target words, among which we include definitions from WordNet and the online Cambridge Dictionary\footnote{Available at \url{https://dictionary.cambridge.org/}} to ensure the presence of highly informative sentences in the manual annotations. 
The final evaluation corpus consists of 7 sentences for each of these 20 target words.


\section{Experiments on test and annotated data}

We extract 20,000 sentences for each of the selected target words: 10,000 containing the target word and 1,000 for each of its 10 distractors. 80\% are used for training, 10\% for development and 10\% for testing. We use a different portion of Gigaword than the one used for the manual annotation. \\
We compare our methods to a simple c2v-based baseline. Given a sentence with a target word, we calculate the similarity between the vectors of all distractors, as well as of the target word, with the c2v context vector of the sentence.
The potential fillers are sorted by similarity and the rank of the target word is used as an indicator of informativeness 
(the more similar, the more informative).

The classifiers described in Section \ref{ssec:classif} are trained on the 
extracted sentences for each selected target word. Results on the test set are found in Table \ref{tab:resultstestmanual} (first column). The c2v neural model gets the highest accuracy among all classifiers, while the FastText classifier has the lowest. They are then used to make predictions on the manually annotated data. We measure the correlation between the probability of belonging to the target word class and the $info3$ score. Results of this correlation are found in Table \ref{tab:resultstestmanual} (second column).
The mean correlation for each classification algorithm is close to zero. Moreover, correlations vary a lot depending on the target word. We conclude that the informativeness measure represented by $info3$ score is not related to the way our classifiers select the most representative sentences for a target word.

Examples of sentences for the target word \textit{tell} can be found in Table \ref{tab:sentClassiExample}. The first two are definitions, assigned high informativeness scores by the annotators; however, the classifier does not recognize the first one as a highly informative sentence. The last sentence is instead classified as a \textit{tell}-sentence with a very high probability even though humans did not find it informative.

\begin{table}[!ht]
\centering
\begin{tabular}{l|c|c}
Classifier & Accuracy on test data  & Spearman's r on manual data \\
\hline
Linguistic-based & 0.890 & -0.195\\
FastText & 0.868 & \textbf{0.101} \\
c2v Logistic regression & 0.927 & -0.179\\
c2v feed-forward NN & \textbf{0.946} & -0.162 \\
c2v baseline  & 0.759 & -0.070 \\
\end{tabular}
\caption{Classifiers' results. The first column shows the average accuracy across words on the classification test set. The second one indicates the Spearman correlation of each classifier's prediction with informativeness annotations.} 
\label{tab:resultstestmanual}
\end{table}

\begin{table}[!h]
\centering
\begin{tabular}{|l|c|c|}
\hline
\textbf{Sentence}                                                                                                         & \textbf{Human score} & \textbf{Classifier score} \\ \hline
To tell is to let something be known. & 4                    & 0.62                      \\ \hline
To tell means express something in words.                                                                   & 5                    & 0.91                      \\ \hline
What can I tell him ?                                                                                            & 1                    & 0.96                      \\ \hline
\end{tabular}
\caption{FastText classifier's probability to belong to the target word's class compared with human annotation for a set of sentences. The target word is the verb "tell".} 
\label{tab:sentClassiExample}
\end{table}

\section{Experiments on word embeddings training}

We concluded in the previous section that the classifiers output a different kind of informativeness than the human annotations. In this section, we test the effect of the classifiers' informativeness on word embeddings training. We use the probabilities outputted by the Fasttext classifier for this task. Independently of the correlations with the human annotation, this classifier's distribution of outputted probability is the least skewed. The others assign very high probabilities for a large portion of sentences, preventing a clear discrimination between informative and not informative sentences.

Following \citet{Baroni-highrisk}, for each target word, we sort the sentences of the test set by probability of belonging to the class of the target word according to the classifier. We select the 250 sentences with lowest and highest probability, and make also a random selection. The target word is replaced by a new token "\textit{target\_word\_new}" in all sentences. We initialize all weights of a word2vec model \citep{mikolov2013distributed} using pre-trained word embeddings.\footnote{Available at \url{https://code.google.com/archive/p/word2vec/}} Unknown words are initialized randomly.
We fine-tune the pre-trained embeddings on each set of sentences. Then, we compute the similarities between the vector of "target\_word\_new" and the pre-trained embedding of the target word (gold standard).

Table \ref{embedding-results} shows the results of this experiment. 
$sim$-$inf$, $sim$-$uninf$ and $sim$-$random$ are the similarities computed on the sets of 250 most informative sentences, 250 least informative and 250 random, respectively; $sim$-$inf\&uninf$ includes the 250 most informative sentences and 250 least informative.
Looking at the mean differences of each column with $sim$-$inf$, we conclude that a low informativeness score of a sentence towards a target word indicates it is less suitable to learn a word embedding; however, sentences with high probability are not necessarily more helpful than the rest. 
Moreover, training embeddings on a set of 500 good and bad sentences gives almost the same quality of embeddings as training on a set of only 250 good sentences.

The second part of Table \ref{embedding-results} compares the similarities of vectors when trained on 200 random sentences ($sim$-$random200$), and when augmenting them with 50 "uninformative" sentences ($sim$-$random$-$uninf$). The former is usually lower when the $sim$-$uninf$ value is low, showing that removing these sentences can improve word representations. However, for words with a high value of $sim$-$uninf$, the value $sim$-$random$-$uninf$ is still higher than $sim$-$random200$. Thus, the mean difference is close to zero.

We investigate the reason behind the large disparities among words for this task. We consider two indicators for each word: its frequency in Google Unigrams \cite{brants2006web} and its polysemy in WordNet \cite{Fellbaum1998}. 
The Spearman correlation between the word frequency and the difference between $sim$-$inf$ and $sim$-$random$ is high and negative (-0.6) with a p-value $<0.05$. Thus, for frequent words, the difference in similarity is low between highly informative sentences and random sentences. On the contrary, in the case of infrequent words, the informative sentences provide better embeddings than the random ones.
The correlations with the other columns of Table \ref{embedding-results} are not significant.
For polysemy, we divide the words into two classes according to the median: the words with less than 11 different senses in WordNet
and the words with 11 or more. No significant correlation is observed with the similarity values.

\begin{table}[!ht]
\centering
\footnotesize
\begin{tabular}{|c|c|c|c|c|l|c|c|}
\cline{1-5} \cline{7-8}
\textbf{}                                                    & \textbf{sim-inf} & \textbf{sim-uninf} & \textbf{sim-inf\&uninf} & \textbf{sim-rand250} &  & \textbf{sim-rand200} & \textbf{sim-rand\&uninf} \\ \cline{1-5} \cline{7-8} 
mean                                                         & 0.567             & 0.393            & 0.611                  & 0.578                 &  & 0.574                 & 0.580                  \\ \cline{1-5} \cline{7-8} 
\begin{tabular}[c]{@{}c@{}}diff with \\ 1st col\end{tabular} & -                 & 0.174            & -0.044                 & -0.011                 &  & -                     & 0.006                  \\ \cline{1-5} \cline{7-8} 
\end{tabular}
\caption{The table on the left shows the mean similarity (for the 20 target words) between the pre-trained vector of the target words and the vector trained from selecting only 250 sentences with highest and lowest probability as well as 250 random sentences according to the Fasttext classifier's output. The second table shows the effect that adding 50 low probability sentences to 200 random sentences has on the mean similarity.}
\label{embedding-results}
\end{table}

\vspace{-0.5cm}

\section{Discussion and future work}

In this study, we have introduced the notion of informativeness and proposed an automatic method to predict it for a given set of words. We have performed a manual annotation of informativeness and used it to evaluate our models. Despite their efficiency in classifying sentences, classifiers do not perform well on the manual dataset, suggesting that what the models are learning is different from our definition of informativeness. 
However, when using the informative and uninformative sentences predicted by the classifier on the task of word embeddings training, we observe that training on uninformative sentences leads in general to lower quality embeddings. Moreover, the average informativeness score of sentences varies a lot depending on the target word: for infrequent target words, informative sentences usually provide a higher gain compared to random sentences.


While not allowing for strong claims about the impact of informativeness on word representations, we believe the results of the present study put forward several interesting questions worth of further research.
First of all, it remains to be seen whether the human concept of informativeness can be of help to NLP applications. Several modifications can be introduced to our algorithms, such as the number of distractors or a variety of source corpora; and annotating a bigger dataset could possibly allow for supervised learning of informativeness.

Another open question is whether embeddings might benefit more from an alternative concept of informativeness. For instance, although definitions are --or should be-- informative for humans, they are not very common in most corpora. For this reason, language representations trained on common corpora, like the ones our classifiers rely on, may find them atypical. 


With a better understanding of informativeness and automatic predictors, we believe that a study of the features that make a sentence informative for a word would be of great theoretical as well as practical interest, possibly allowing to build target-word-independent predictors. Such features could involve context words sharing a topic with the target, or the degree of polysemy of context words. 

\newpage



\bibliographystyle{taln2019}
\bibliography{biblio}
\nocite{}

\end{document}